\pdfoutput=1
\documentclass{article} 
\usepackage{nips15submit_e,times}
\usepackage{url}
\usepackage[utf8]{inputenc}
\usepackage[OT1]{fontenc}
\usepackage[finnish,swedish,english]{babel}
\usepackage{pbox}
\usepackage{csquotes}
\usepackage{eurosym} 
\usepackage{multirow}
\usepackage{amssymb}
\usepackage{amsmath}
\usepackage{verbatim}
\usepackage{longtable}
\usepackage{subfigure}
\usepackage[medium]{titlesec}
\usepackage{tikz}
\usetikzlibrary{positioning}
\usetikzlibrary{calc}
\usetikzlibrary{arrows}
\usetikzlibrary{decorations.pathmorphing,decorations.markings}
\usetikzlibrary{shapes}
\usetikzlibrary{patterns}
\usepackage{pgfplots,pgfplotstable}
\usepackage{graphicx}

\title{A Character-Word Compositional Neural Language Model for Finnish}

\author{
Matti Lankinen \\
Reaktor \\
\texttt{matti.lankinen@reaktor.com} \\
\And
Hannes Heikinheimo \\
Reaktor \\
\texttt{hannes.heikinheimo@reaktor.com} \\
\AND
Pyry Takala \\
Aalto University \\
\texttt{pyry.takala@aalto.fi} \\
\And
Tapani Raiko \\
Aalto University  \\
\texttt{tapani.raiko@aalto.fi} \\
\And
Juha Karhunen \\
Aalto University \\
\texttt{juha.karhunen@aalto.fi} \\
}

\nipsfinalcopy 

\begin{document}

\maketitle

\begin{abstract}
Inspired by recent research, we explore ways to model the highly morphological Finnish language at the level of characters while maintaining the performance of word-level models. We propose a new Character-to-Word-to-Character (C2W2C) compositional language model that uses characters as input and output while still internally processing word level embeddings. Our preliminary experiments, using the Finnish Europarl V7 corpus, indicate that C2W2C can respond well to the challenges of morphologically rich languages such as high out of vocabulary rates, the prediction of novel words, and growing vocabulary size. Notably, the model is able to correctly score inflectional forms that are not present in the training data and sample grammatically and semantically correct Finnish sentences character by character.   
\end{abstract}

\section{Introduction}\label{chapter:intro}
Character level language models have attracted significant research attention recently in the deep learning community \cite{lee2016fully,cho2016,ling2015,bojanowski2015,google2016,kim2015}. Character level language models have certain advantages over word level models, which tend to struggle with applications that require a large vocabulary or need to tackle high out-of-vocabulary word rates. In many cases increasing vocabulary size quickly multiplies the parameter count, making word-level models complex and slow to train. Furhermore, with highly inflectional languages such as Turkish, Russian, Hungarian or Finnish word level models fail to encode part-of-speech information embedded into the morphology of inflected words. 

 Among highly morphological languages Finnish is one of the most complex. Hence the Finnish language provids an interesting testbed for research in character level language models. It has a large number of inflectional types for both nouns and verbs and it uses suffixes to express grammatical relations. As an example, the single Finnish word \textit{talossanikin} corresponds to the English phrase \textit{in my house, too}. Bojanowski et al. \cite{bojanowski2015} report the vocabulary size and out-of-vocabulary rate of various European languages for a set of experiments on the Europarl dataset \cite{koehn2005europarl}. For Finnish these values were reported to be 10 \% and 37 \% higher respectively, compared to Hungarian, which had the second hightest reported values.

\begin{figure}[h]
\centering \includegraphics[width=0.6\textwidth]{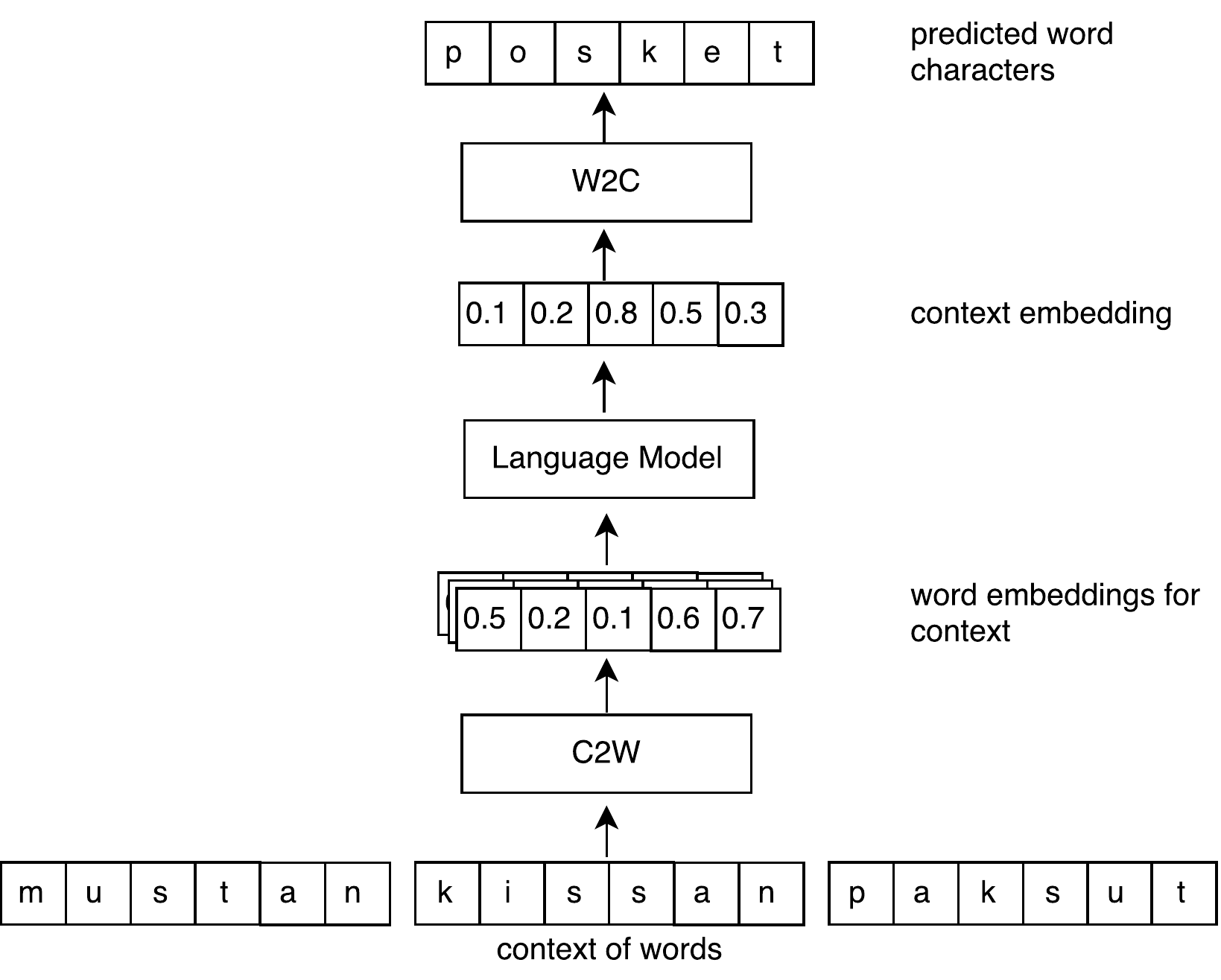} \caption{C2W2C model overall structure} \label{fig:c2wc2_stucture}
\end{figure}

In this study we propose a new \textbf{Character-to-Word-to-Character} (C2W2C) compositional language model for modeling Finnish. The model combines ideas from \cite{ling2015} and \cite{cho2016}, and consists of three logical parts: (1) the Character-to-Word C2W model inspired by \cite{ling2015}, (2) a traditional LSTM Language Model \cite{sundermeyer2012lstm}, and (3) a Word-to-Character W2C decoder model inspired by \cite{cho2016}. The entire model is implemented sequentially, so the output from any logical part can be used as an input for the next one. Thus, the model can be trained and used with a single pass without any preliminary sub-word tokenization or word classification. Also, all the parts are independent, so they can be trained individually. The structure of C2W2C model is described in Figure \ref{fig:c2wc2_stucture}.

The rest of our discussion is organised as follows: Section~\ref{chapter:background} discusses related work. Section~\ref{chapter:c2w2c} describes our language model architecture in more detail. Section~\ref{chapter:experiments} and \ref{chapter:results} present our experiment  setup and the model performance results. Section~\ref{chapter:conclusions} contains the conclusions.

\section{Related work} \label{chapter:background}

The usual approach in neural language modeling is to use so-called one-hot vectors \cite{money2007} to represent input and output words of the model. However, the number of parameters linearly follow the size of the vocabulary, hence resulting into scalability issues for large vocabularies. Ling et .al \cite{ling2015} proposed a new Character-to-Word model (C2W) to solve this issue. In C2W, the traditional word-level projection layer is replaced with a character-level Bidirectional LSTM unit that produced word embeddings to the language model LSTM. As a result, the input layer parameters are reduced from 4M to 180k with a dataset from the Wikipedia corpus. Also noticeable performance improvements were reported compared to traditional word-level models. Here we use the C2W model of \cite{ling2015} as part of our C2W2C model. 

In \cite{kim2015} Kim et al. propose a convolutional neural network  and a highway network over characters (CharCNN), whose output is given to a LSTM language model. With this setup, the authors managed in par performance with state-of-art models with 60\% fewer parameters.  The work in \cite{google2016} proposes also a solution by using convolution networks, but use a CharCNN as a part of the output layer. The architecture of this model followes the original work of \cite{kim2015}, but instead of having the projection layer for output, they use a CharCNN to create word embeddings, which they combine  with a context vector from a language model layer to produce an output word logit function.  Other alternative approaches in dealing with the large output vocabulary size are the hierachical softmax \cite{goodman2001classes,mnih2009scalable}, importance sampling \cite{bengio2008adaptive}, and noice contrast estimation \cite{gutmann2010noise,mnih2013learning}. Although these methods work well and give speedups without significant performance penalties, they do not remove the challenges of out-of-vocabulary words.

\begin{figure}[h]
\centering 
\includegraphics[width=0.7\textwidth]{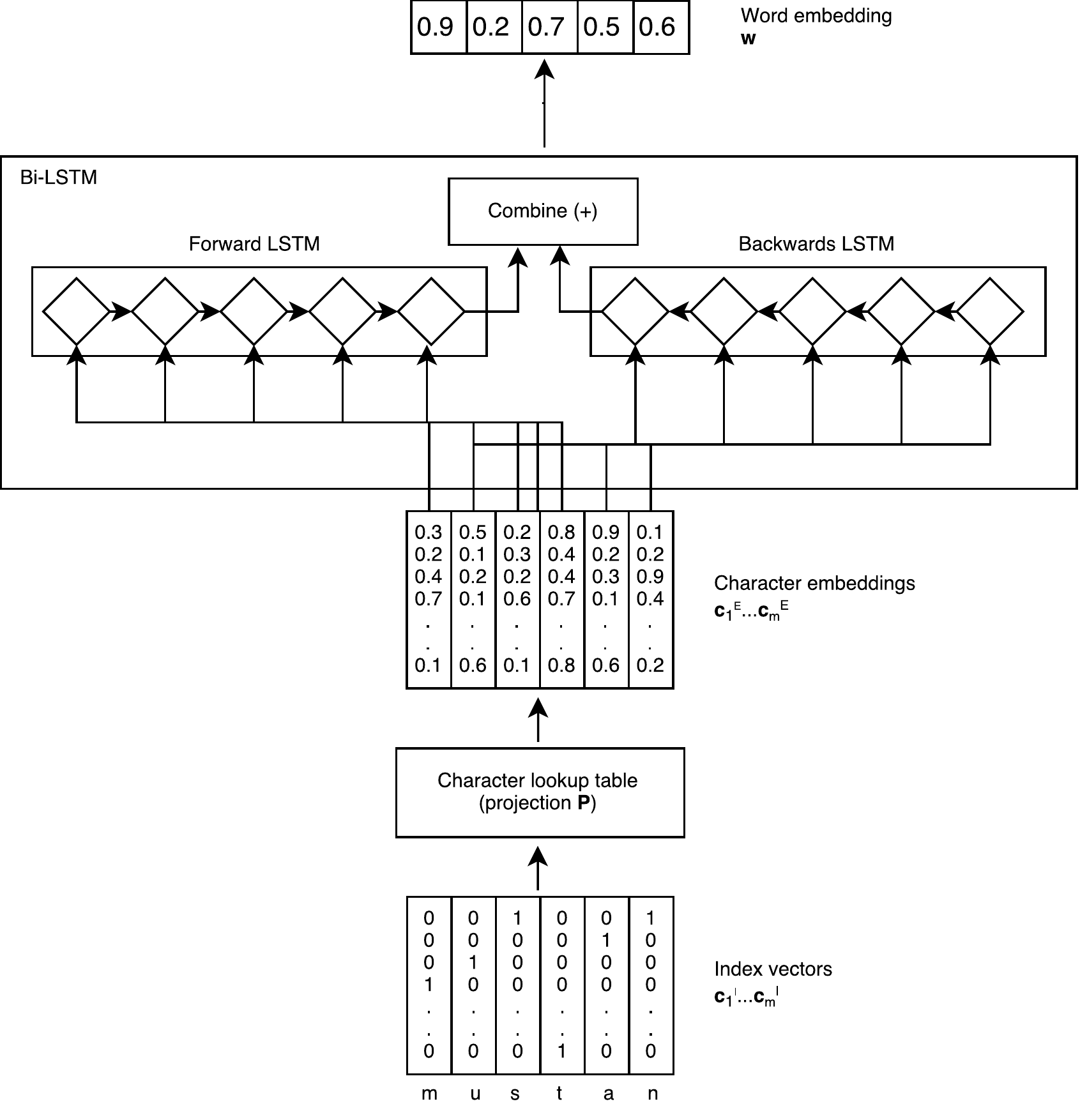} 
\caption{The structure of C2W model}
\label{fig:c2w}
\end{figure}

Cho et al. \cite{cho2016} proposed a new two-stage encoder-decoder RNN performing machine translations from characters to characters. They use the previous encoder-decoder work of \cite{cho2014} and gated-feedback networks \cite{chung2015} as a basis and built an adaptive "Bi-Scale Recurrent Neural Network", where the idea is to model sentences by using two layers. The "faster layer" models the fast-changing timescale (i.e. words), and the "slower layer" modeles the slower-changing timescale (i.e. characters). The layers were gated so that the produced context vector was an adaptive combination of the activations of both layers, and that context was used to decode the translation characters. In this paper we use a character-level decoder as a part of its C2W2C model similar to \cite{cho2016}.

Recent research, including \cite{mikolov2012,bojanowski2015,luong2013} have indicated the need for more research on subword level models for morphologically rich languages. However, sub-word level language models have been studies previously. Creutz and Lagus \cite{creutz2005unsupervised} introduced an unsupervised morpheme segmentation and morphology induction algorithm called Morfessor. Vilar et al. \cite{vilar2007} translated sentences character by character, but the results of this technique were inferior to those of word-based approaches. Botha and Blunsom \cite{botha2014} proposed a word segment-level language model. 

In this paper we apply neural language modelling especilly to the Finnish language. Previous work done towards tackling the complexites of Finnish morpoholgy include \cite{linden2011hfst} and \cite{silfverberg2015finnpos}. Most recent work on neural language modeling applied to the Finnish language are \cite{ginter2014fast} and \cite{enarvi2016theanolm}.  Chung et al. \cite{cho2016} demonstrate their machine translation work using Finnish as one of their target languages.

\section{A compositional language model} \label{chapter:c2w2c}

\subsection{Character to Word model}\label{subsectionC2W}

\begin{figure}[h]
\centering \includegraphics[width=0.7\textwidth]{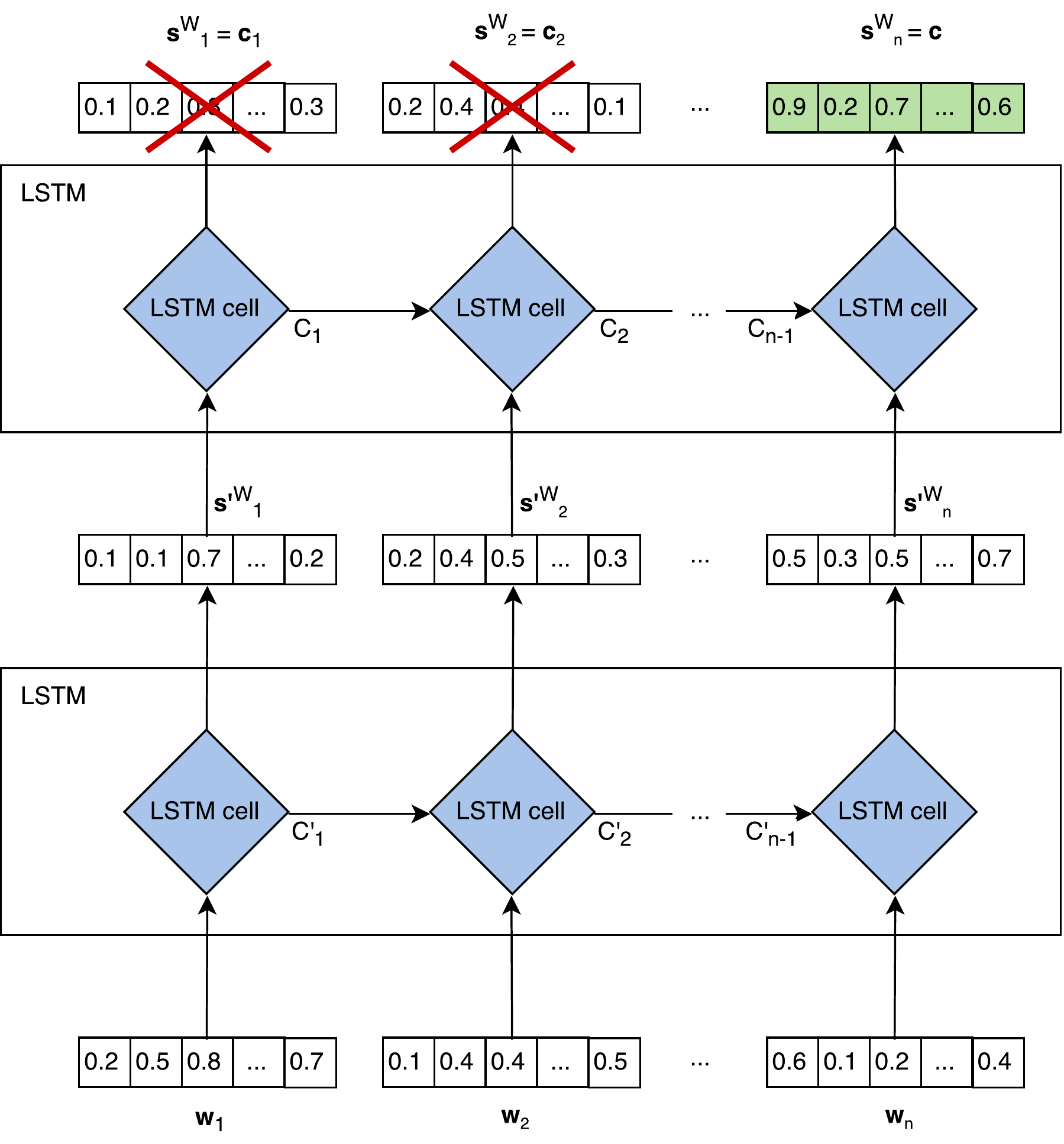} \caption{The C2W2C language model structure} \label{fig:lm_stucture}
\end{figure}

The character-to-word (C2W) model is the first part of the C2W2C model. The implementation mainly follows \cite{ling2015}. The responsibility of C2W in C2W2C is to transform the input context word sequence \(\omega_1...\omega_n\) into \(d_W\)-dimensional \emph{word embeddings} \(\mathbf{w}_1...\mathbf{w}_n\). The word embeddings capture the characteristics of each of the words in the given context. These characteristics contain, for example, part-of-speech information (noun, verb, adjective, etc.), grammatical tense, and inflection information that is usually encoded into the words in highly inflectional languages (e.g. Finnish words \emph{auto, autossa, autosta} as opposed to their English counterparts \emph{a car, in the car, from the car}).

C2W2C contains the first two layers from the original C2W model implementation: (1) character lookup table and (2) Bi-LSTM generating the word embeddings for the given word sequence \(\omega_1...\omega_n\). These first two C2W layers are illustrated in Figure \ref{fig:c2w}. More details can be found from \cite{ling2015}.

\subsection{Language Model}\label{subsectionlanguagemodel}

The language model component of C2W2C is a standard 2-layer LSTM network \cite{hochreiter1997} that takes word embeddings \(\mathbf{w}_1...\mathbf{w}_n\) as an input and yields the state sequence \(\mathbf{s}^W_0...\mathbf{s}^W_n\) as an ouput. In training time, the intermediate states \(\mathbf{s}^W_0...\mathbf{s}^W_{n-1}\) are discarded and the actual \emph{context embedding} \(\mathbf{c}\) is selected by taking the last state of the computed state sequence. The overall structure of the LM of C2W2C is shown in Figure \ref{fig:lm_stucture}.

\subsection{Word to Character model}

The Word to Character (W2C) model is the final part of C2W2C mode. Its goal is to decompose the context embedding \(\mathbf{c}\) back to a sequence of characters along the lines of the work of \cite{cho2016} and \cite{cho2014}.

The full implementation of the character-level encoder-decoder \cite{cho2016} contains two adaptive decoder-RNN \cite{cho2014} layers whose goal is to translate characters from a source word sequence \(\mathbf{x}_1...\mathbf{x}_n\) into a target word sequence \(\mathbf{y}_1...\mathbf{y}_n\). However, since C2W2C tries to predict only a single target word, our W2C implementation has only one layer following the details of the original decoder-RNN.

\begin{figure}[h]
\centering \includegraphics[width=0.7\textwidth]{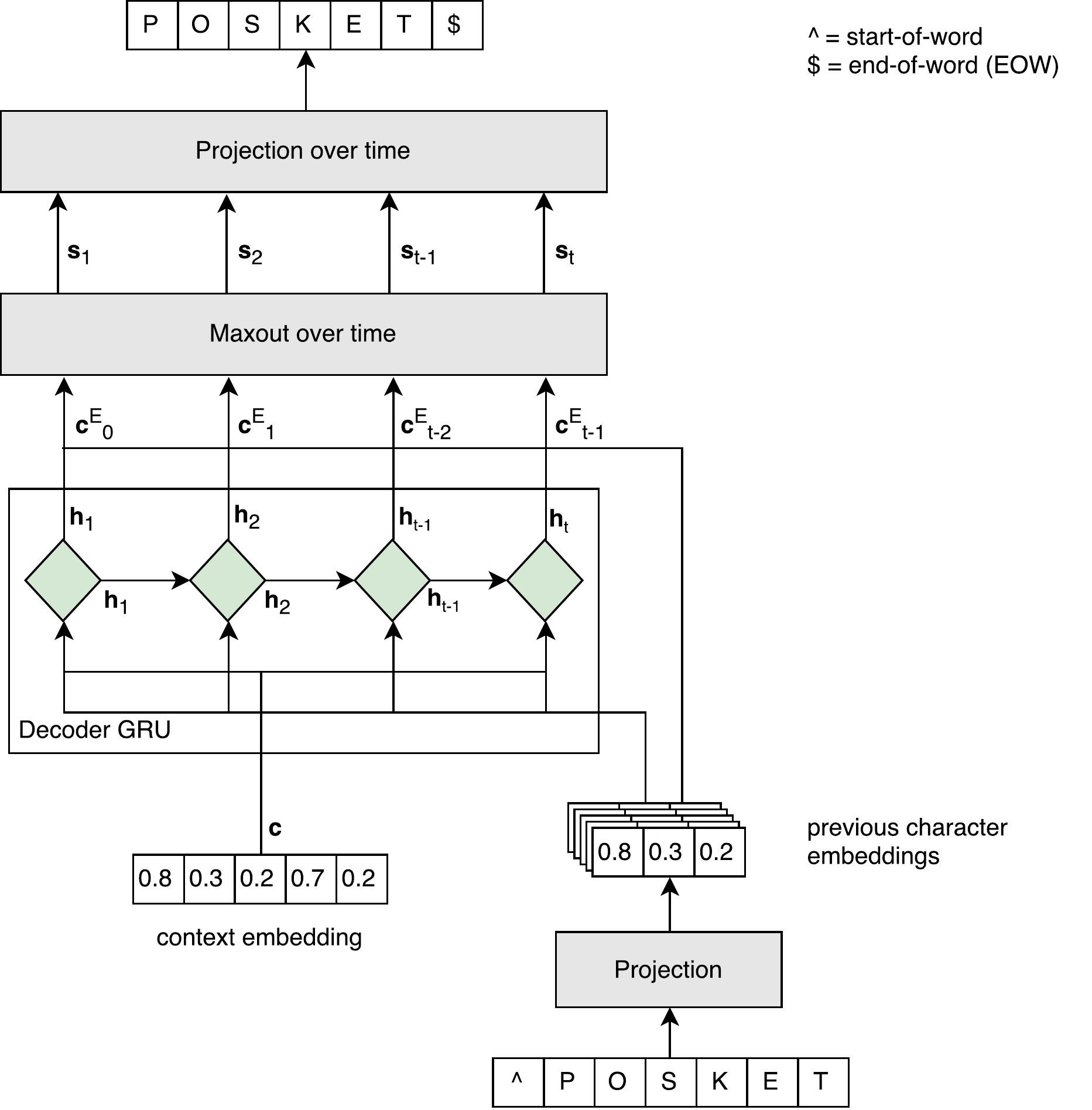} \caption{The structure of W2C model} \label{fig:w2c}
\end{figure}

W2C works as follows: The layer receives the context embedding \(\mathbf{c}\) from the language model and combines it with the predecessor character \(c_{t-1}\) of the predicted character \(c_i\). By using these two components, the RNN decoder maintains a hidden state \(\mathbf{h}\) over every predicted character. The used RNN implementation is a variant of Gated Recurrent Unit \cite{cho2014}, with the difference that instead of relying only on the hidden state \(\mathbf{h}_{t-1}\), the RNN also uses an embedding of the predecessor character \(\mathbf{c}^E_{t-1}\) to get the next state \(\mathbf{h}_t\) by using the following formula:

\begin{equation} 
\begin{split} 
\mathbf{h}_t    & = z_t \mathbf{h}_{t-1} + (1 - z_t) \mathbf{h'}_t \\ 
\mathbf{h'}_t   & = \tanh ( \mathbf{W}_h \mathbf{c}^E_{t-1} + r_t(\mathbf{U}_h\mathbf{h}_{t-1} + \mathbf{C}_h \mathbf{c}) ) \\ 
z_t             & = \sigma ( \mathbf{W}_z \mathbf{c}^E_{t-1} + \mathbf{U}_z\mathbf{h}_{t-1} + \mathbf{C}_z \mathbf{c}) \\ 
r_t             & = \sigma ( \mathbf{W}_r \mathbf{c}^E_{t-1} + \mathbf{U}_r\mathbf{h}_{t-1} + \mathbf{C}_r \mathbf{c}) 
\end{split} 
\end{equation}

The initial \(\mathbf{c}^E_0\) is a zero vector, and the initial hidden state is initialized by using the context vector and \(\tanh\) activation:

\[ \mathbf{h}_0 = \tanh(\mathbf{V}\mathbf{c}) \]

Each state \(\mathbf{h}_t\) of the received state sequence \(\mathbf{h}_1...\mathbf{h}_m\) is combined with the context vector and the previous character embedding \(\mathbf{c}^E_{t-1}\), and the logit for the predicted character index (in \(V_C\)) is received by running the combined embedding through a 2-feature Maxout network \cite{goodfellow2013} and projecting the result to the \(V_C\) index space:

\begin{equation} 
\begin{split} 
\mathbf{c}^I_t   & = \mathbf{P}_I \mathbf{s}_t \\ 
\mathbf{s}_t     & = \max \{ \mathbf{s'}^1_t , \mathbf{s'}^2_t \} \\ 
\mathbf{s'}^i_t  & = \mathbf{O}^i_h \mathbf{h}_t + \mathbf{O}^i_e \mathbf{c}^E_{t-1} + \mathbf{O}^i_c\mathbf{c} + \mathbf{b} 
\end{split} 
\end{equation}

Probabilities of the predicted characters are received by running Softmax activation over the logit vectors. The overall structure of W2C is illustrated in Figure \ref{fig:w2c}.
 
\section{Experiment: Finnish Language Modeling} \label{chapter:experiments}

We examine how well the C2W2C model can model the Finnish language. We are interested in the models capability to learn word-level relationships, find the meaning of the different inflection forms  (e.g. \emph{auton}, `of the car', \emph{autossa}, `in the car') as well as cope with part of speech information (e.g. \emph{Pekka ajaa punaista autoa}, `Pekka drives a red car') of a word in the context of a sentence.

\subsection{Dataset} \label{sect:exp:dataset}

We chose the text corpus of Finnish translations of Europarl Parallel Corpus V7\footnote{http://www.statmt.org/europarl}  \cite{koehn2005europarl} as our test dataset. The text data was was tokenized with Apache OpenNLP open-source tool\footnote{https://opennlp.apache.org} and using Finnish tokenizing model from Finnish Dependency Parser project\footnote{https://github.com/TurkuNLP/Finnish-dep-parser}. The tokenized text data was converted into a lower-case form. No further pre-processing or word segmentation was done to the dataset.

The training set was selected by using the first one million tokens from the tokenized lower-case corpus. The properties of the selected dataset reflected the nature of highly inflectional languages; the one million tokens included \emph{88,000} unique tokens which cover 8.8\% from the dataset. 5,000 most frequent tokens covered 78.41 \% from the dataset, 10k most frequent tokens 84.97 \% and 20k most common frequent tokens 90.55 \%. 

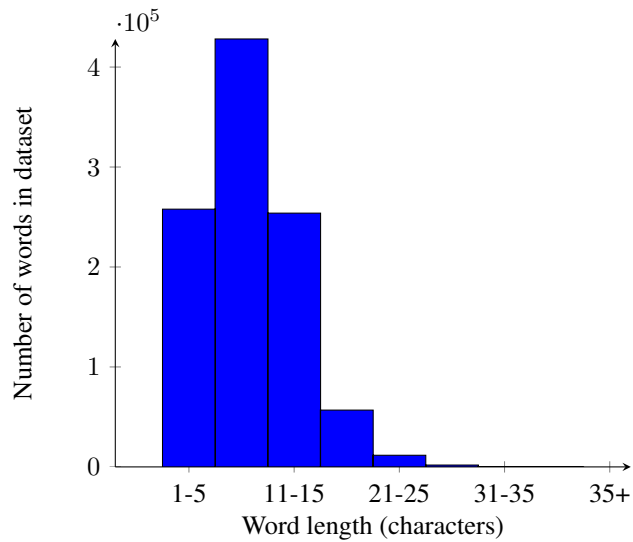
\begin{figure}[h] \centering \begin{tikzpicture}
\begin{axis}[
ybar, bar width=20pt, xlabel={Word length (characters)}, ylabel={Number of words in dataset}, ymin=0, axis x line=bottom, axis y line=left, enlarge x limits=0.2, symbolic x coords={1-5, 6-10, 11-15, 16-20, 21-25, 26-30, 31-35, 35+}, xticklabel style={anchor=base,yshift=-\baselineskip}]
\addplot[ybar,fill=blue] coordinates {
(1-5,257842) (6-10,428108) (11-15,253875) (16-20,56828) (21-25,11648) (26-30,1851) (31-35,190) (35+,25)
}; \end{axis}
\end{tikzpicture} \caption{Number of words in dataset by word length} \label{fig:words_in_dataset_histo} \end{figure}

In addition to the training data, extra 10k tokens were fetched from the same Europarl Finnish corpus for test and validation purposes. Both training and validation datasets were pre-processed as described above and sentences were split per line and shuffled. The training dataset sentences were also shuffled at the beginning of each epoch.

\subsection{Hyper-parameters}

The hyper-parameters of C2W model were set to follow the values from the original C2W proposed by \cite{ling2015}. The projection layer \(\mathbf{P}_E \in \mathbb{R}^{d_C\times|V_C|}\) for the character embeddings was set to use character properties \(d_C = 50\). The intermediate Bi-LSTM cell states \(\mathbf{s}^f_m\) and \(\mathbf{s}^b_0\) were both set to use \(d_{WI} = |\mathbf{s}^f_m| = |\mathbf{s}^b_0| = 150\). Word features number \(d_W\) was set to 50. The language model LSTM hidden state \(d_L\) was set to 500 for both nodes. W2C decoder's hidden state was set to \(|\mathbf{h}| = 500\) and character embeddings for both \(c^E_{t-1}\) and \(\mathbf{s}_t\) were set to use \(d_C\).

Another variable that has a great influence on the model speed is the maximum length of the input and output words: the longer the maximum length, the more inner LSTM steps must be performed to construct the word embeddings. Fig.~\ref{fig:words_in_dataset_histo} shows the word frequency distribution by word length. We chose the maximum word length to be 20, which covers over 98\% from the training dataset.

The training was performed by using Backpropagation through time with unlimited context size and Adam optimizer \cite{kingma14} with learning rate of 0.0001and norm clipping with value 2.0. Gradient updates were performed after each time step, and language model's LSTM states were reset before each epoch. In order to speed up the calculation, the data was arranged into batches so that each sample \(x^t_i\) in the batch was a predecessor of the next sample \(x^{t+1}_i\). Size of these mini-batches were set to 150. To prevent overfitting we used dropout layers \cite{srivastava2014} after each submodel (W2C, LM, W2C) with friction value 0.5. 

\vspace{-0.25cm}
\subsection{Perplexity} 

For character-level perplexity the probability of a word is seen as a product of the probabilities of its characters. Because C2W2C already produces categorical probability distributions by using softmax as its output layer activation, the perplexity for C2W2C was calculated by using the received probability distribution and getting the loss character by character. Those character losses are then added per word, and finally, the word-level cross-entropies are combined and normalised with the testing set size to get overall validation perplexity.

\vspace{-0.20cm}
\subsection{Environment}

The model was trained and validated by using a single Ubuntu 14.04 server having 8GB RAM, single Intel i5 for the model building, and one nVidia GeForce GTX 970 GPU Unit with 3.5GB graphics memory and CUDA (cMEM and cuDNN disabled) for the model training. 

The model was implemented by using Python 2.7, Theano 0.8.2\footnote{http://deeplearning.net/software/theano} and Keras 1.0.4\footnote{http://keras.io}. The source codes of the final C2W2C model implementation can be found at GitHub: \url{https://github.com/milankinen/c2w2c} under MIT license.
 
\section{Results and Analysis} \label{chapter:results}

The model was run by using the dataset, hyperparameters, and environment described in the previous section. In addition to C2W2C, a traditional word-level model ("Word-LSTM" onwards) was trained for comparison. Word-LSTM had the same two-level LSTM language model as C2W2C (as described in subsection~\ref{subsectionlanguagemodel}) and typical feedforward projection layers for both inputs and outputs. For the sake of efficiency, context vector \(\mathbf{c} |_{dim(\mathbf{c}) = 500}\) was projected into 150 dimensional space before output projection (removing ~31M parameters from the model).

\subsection{Quantitative analysis}

\subsubsection{Model Complexity and training times} 

As shown in Table \ref{table:model_params} the C2W2C model operates in our experiment by using only 25\% of the traditional word-level model parameters. The C2W layer, in particular, seems very useful as it reduces the input parameters by a factor of 20 without affecting model performance negatively.

\begin{table}[h] \centering \begin{tabular}{|c|c|c|c|c|c|} \hline \multicolumn{3}{|c|}{\textbf{C2W2C}}      & \multicolumn{3}{c|}{\textbf{Word-LSTM}}       \\ \hline \textbf{C2W} & \textbf{LM} & \textbf{W2C} & \textbf{FF-NN} & \textbf{LM} & \textbf{FF-NN} \\ \hline 0.26M          & 3.1M         & 2.04M          & 4.5M            & 3.1M         & 13.2M         \\ \hline \multicolumn{3}{|c|}{5.41M}                & \multicolumn{3}{c|}{20.8M}                     \\ \hline \end{tabular} \caption{Model parameters per sub-model} \label{table:model_params} \end{table}

However, even if there are significantly fewer parameters, the training times are roughly equal: Both models can process approximately 900 words per second, resulting in 20 minutes per epoch with the given training data set. However, for the C2W2C the gradient is less stable compared to the World-LSTM model. To remendy this we had to resort to a smaller learning rate, hence requiring more iterations for similar results.  

\subsubsection{Perplexity} 

When measured with perplexity, the classic word-level model surpasses C2W2C. With the given dataset, word-level model achieved PP 392.28, whereas PP for C2W2C was 410.95. One possible factor that raises the PP of C2W2C is that it gives predictions to \emph{all} possible character sequence, which means \(|V_C|^{maxlen}\) different combinations. However, the predicted vocabulary size is much smaller. The work done in \cite{google2016} eliminated this by doing an expensive normalisation over the training vocabulary, which indeed improved character-level performance a little bit. However, such normalisation was not done in the scope of this study.

\subsection{Qualitative analysis}

\subsubsection{Grammar} 

The following table displays some example scores (a length-normalized log-loss of the sentence, where a lower score is better) of sentences with different properties. The bolded word \textbf{w} in the first column represents the examined word and the second column lists the tested variants for \textbf{w}. None of the listed sentences is found from the training dataset.

The results indicate that C2W2C can learn the stucture of Finnish grammar and capture morphological information that is encoded into words. Especially the sense of the word (e.g. \emph{esittää, esitti, on esittänyt}'to present, presented, has presented' ) is a property where correct words yield better scores than incorrect ones. Same applies to inflection forms (e.g. väittei\emph{stä}, väittei\emph{ltä}, väittei\emph{lle}, `from the arguments', `of the arguments', `for the arguments') especially when there are more than two subsequent words having same inflection form.

However, C2W2C is less good at predicting dual forms (whether a word is in singular or plural form) | usually the model prefers singular forms. One possible reason for this might be the score function; word score in C2W2C is not length-normalized, which prefers shorter singular words (e.g. \emph{talo} versus \emph{talot} `house' vesus `houses', \emph{äänestyksen} versus \emph{äänestyksien} `vote' versus `votes').

Perhaps the most interesting feature of the C2W2C model is how well it can predict words that are not part of the dataset. Whereas traditional word-level models would use an unknown token for such words, C2W2C can give scores to the actual words as long as they do not contain any unknown characters. For example, even if words \emph{sijoitusrahasto} `mutual Fund' and \emph{monivuotisella} `multiannual' in Table \ref{table:res:inflection} do not belong to the training data, C2W2C can still assign a better scores to the grammatical correct unseen wordform than to their morphological variants found from the dataset!

\begin{table}[h] \small \centering \begin{tabular}{|l|l|l|} \hline \textbf{Sentence} & \textbf{Variants} & \textbf{Score} \\ \hline \multirow{2}{*}{\textless S\textgreater \space Olen huolestunut esitetyistä \textbf{w} . \textless/S\textgreater} & väitteistä* & \textbf{5.143} \\ \cline{2-3}
& väite & 6.315 \\ \hline \multirow{2}{*}{\textless S\textgreater \space Komission on \textbf{w} aloitteen . \textless/S\textgreater} & esittänyt* & \textbf{2.972}  \\ \cline{2-3}
& esittää & 4.254 \\ \hline \multirow{2}{*}{\textless S\textgreater \space Nämä esitykset vaikuttavat \textbf{w} . \textless/S\textgreater} & järkeviltä* & 5.557 \\ \cline{2-3}
& järkevältä & \textbf{5.045} \\ \hline \multirow{2}{*}{\textless S\textgreater \space Tämä \textbf{w} vaikuttaa kannattavalta . \textless/S\textgreater} & sijoitusrahasto* & \textbf{6.503} \\ \cline{2-3}
& sijoitusrahastot & 7.605 \\ \hline \multirow{3}{*}{\pbox{20cm}{\textless S\textgreater \space Säästöt aiotaan saavuttaa tällä \textbf{w} \\ ohjelmalla . \textless/S\textgreater}} & monivuotisella* & \textbf{6.503} \\ \cline{2-3}
& monivuotista & 7.658 \\ \cline{2-3} & monivuotinen & 7.605 \\ \hline
\end{tabular} \caption{Word variant scores (lower is better) with different inflection forms. The grammatically correct alternative marked with *.} \label{table:res:inflection} \end{table}

\subsubsection{Semantics} 

Results indicate that C2W2C can learn semantic meanings and part-of-speech information | for example, if a verb is replaced with a noun, it usually yields worse scores than the correct sentence. The same applies to semantics: for example, entities that can do something usually yield better scores when placed before verbs than passive entities in the same position. However, punctuation seems to be harder for C2W2C. Probably due to a significant number of statements and the nature of the dataset the model typically assigns better scores to periods than question marks, even if the sentence starts with a question word. 

A few example sentences and their scores are shown in Table \ref{table:res:semantics}. The notation and used scoring are same as in Table \ref{table:res:inflection}.

\subsection{Sampling text with C2W2C}

As an example application, the trained C2W2C model was configured to generate the Finnish political text character by character. Two different strategies were tested: stochastic sampling and beam search sampling. Stochastic sampling is a simple approach where given a context of words, the next most likely word is chosen and used as a part of the context when predicting the next character. Beam search is a heuristic search algorithm that holds \(k\) most likely word sequences and expands them until all sequences are terminated (\textless/S\textgreater  \space is encountered).

Within those two strategies, beam search gave significantly better results, whereas stochastic sampling usually ended up looping certain phrases. Beam search also had some phrases and words it preferred over others, but the generated samples were much more natural, and the inflection of words was better than with stochastic sampling. A two-layer search beam was used for the text sampling. Individual words were sampled with the beam of \(k=20\), and the sampled words and their probabilities were used with the sentence-level beam of \(k=10\).

The C2W2C model can produce, grammatically and semantically, sensible text, character by character. Some of the best example sentences are displayed below (capitalization was added, and extra spaces were removed afterwards). The first two words of each sentence were given as an initial context, and the rest of the words were sampled by the model.

\begin{displayquote}
\textless S\textgreater \space Haluan kiittää sydämellisesti puheenjohtajavaltio Portugalia hänen mietinnöstään. \footnote{I would like to cordially thank the Portuguese Presidency for his report.} \textless/S\textgreater

\textless S\textgreater \space Nämä ovat erittäin tärkeitä eurooppalaisia mahdollisuuksia Euroopan Unionin jäsenvaltioiden perustamissopimuksen yhteydessä.\footnote{These are very important European opportunities in the context of the Treaty of the European Union Member States.} \textless/S\textgreater

\textless S\textgreater \space Siitä huolimatta haluaisin onnitella esittelijää hänen erinomaisesta mietinnöstään. \footnote{Nevertheless I would like to congratulate the rapporteur on his excellent report.}\textless/S\textgreater

\textless S\textgreater \space Tämä merkitsee sitä, että Euroopan Unionin perustamissopimuksen voimaantulon tarkoituksena on kuitenkin välttämätöntä eurooppalaisille jäsenvaltioille.\footnote{This means that the entry into force of the Treaty of the European Union's aim is, however, indispensable for the European Member States.} \textless/S\textgreater \end{displayquote}

\begin{table}[h] \small \centering \begin{tabular}{|l|l|l|} \hline \textbf{Sentence} & \textbf{Variants} & \textbf{Score} \\ \hline \multirow{3}{*}{\textless S\textgreater \space \textbf{w} on pyytänyt uutta äänestystä . \textless/S\textgreater} & Belgia* & \textbf{5.160} \\ \cline{2-3}
& elokuu & 6.963 \\ \cline{2-3} & ilmiö & 5.345 \\ \hline \multirow{3}{*}{\textless S\textgreater \space Parlamentti \textbf{w} yksimielisesti asetuksesta . \textless/S\textgreater} & päätti* & \textbf{5.095} \\ \cline{2-3} & komissio & 7.339 \\ \cline{2-3} & suuri & 6.628 \\ \hline
\multirow{3}{*}{\textless S\textgreater \space Oletteko tietoisia , että uhka on ohi \textbf{w}  \textless/S\textgreater} & ?* & 5.455 \\ \cline{2-3}
& . & \textbf{5.164} \\ \cline{2-3} & , & 6.154 \\ \hline
\end{tabular} \caption{Sentence scores (lower is better) with different semantical variants. The grammatically correct alternative marked with *.} \label{table:res:semantics} \end{table}

\section{Conclusions} \label{chapter:conclusions}

We propose a new C2W2C character-to-word-to-character compositional language model and estimate its performance compared to the traditional word-level LSTM model. The experiments and their validations are done using Finnish language and the Europarl V7 corpus.

Our experiments are sill preliminary, however, the discoveries mainly follow the findings of \cite{google2016} and \cite{cho2016} | it is indeed possible to model succesfully a morphologically rich  language such as Finnish character by character and to significantly reduce model parameters, especially with large corpuses. Larger scale experiments still need to be conducted, including other data sets and languages, to make further conclusions from the results. 

With our setup the traditional word-level model still slightly outperforms the C2W2C model in terms of perplexity and convergence speed, but the C2W2C model compensates by successfully handling out-of-vocabulary words. The model can also learn the Finnish grammar and assign better scores to grammatically correct sentences than incorrect ones, even though the inflection forms in the sentence are not present in training data. The same applies to text sampling: C2W2C can sample grammatically and semantically correct Finnish sentences character by character.

\bibliography{master}
\bibliographystyle{plain}

\end{document}